\documentclass[twocolumn]{mc_nankai}

\usepackage{colortbl}
\usepackage{utfsym}
\usepackage{booktabs}
\usepackage{amsmath, amssymb, mathtools, bm}
\usepackage{multirow, enumitem, subcaption, float, wrapfig}
\usepackage{pifont}
\usepackage{overpic}

\def\ours{OffSeg}
\usepackage{pifont}
\newcommand{\cmark}{\ding{51}}
\newcommand{\xmark}{\text{\ding{55}}}

\definecolor{iccvblue}{rgb}{0.21,0.49,0.74}

\title{Revisiting Efficient Semantic Segmentation: Learning Offsets for Better Spatial and Class Feature Alignment}

\author[1]{Shi-Chen Zhang}
\author[1]{Yunheng Li}
\author[2]{Yu-Huan Wu}
\author[1]{Qibin Hou$^{\dagger}$}
\author[1]{Ming-Ming Cheng}

\affiliation[1]{VCIP, CS, Nankai University}
\affiliation[2]{IHPC, A*STAR, Singapore}
\affiliationenter[]{$^{\dagger}$Corresponding author.}

\abstract{
Semantic segmentation is fundamental to vision systems requiring pixel-level scene understanding, yet deploying it on resource-constrained devices demands efficient architectures. 
Although existing methods achieve real-time inference through lightweight designs, we reveal their inherent limitation:
misalignment between class representations and image features caused by a per-pixel classification paradigm. 
With experimental analysis, we find that this paradigm results in a highly challenging assumption for efficient scenarios: Image pixel features should not vary for the same category in different images.
To address this dilemma, we propose a coupled dual-branch offset learning paradigm that explicitly learns feature and class offsets to dynamically refine both class representations and spatial image features. 
Based on the proposed paradigm, we construct an efficient semantic segmentation network,~\ours{}.
Notably, the offset learning paradigm can be adopted to existing methods with no additional architectural changes.
Extensive experiments on four datasets, including ADE20K, Cityscapes, COCO-Stuff-164K, and Pascal Context, demonstrate consistent improvements with negligible parameters.
For instance, on the ADE20K dataset, our proposed offset learning paradigm improves SegFormer-B0, SegNeXt-T, and Mask2Former-Tiny by 2.7\%, 1.9\%, and 2.6\% mIoU, respectively, with only 0.1-0.2M additional parameters required.
}

\mclink[Project Page]{\url{https://github.com/HVision-NKU/OffSeg}}

\begin{document}
\maketitle
\justifying

\section{Introduction}
\label{sec:intro}

\begin{table*}
\centering
\small
\caption{Comparison of different semantic segmentation paradigms.
`Fea.' and `Rep.' donate feature adaptation and class representation adaptation, respectively.}
\begin{tabular}{l|ccccc}
\toprule
Paradigm              & Fea. & Rep. & Interaction & Alignment & Overhead \\ \midrule
Per-Pixel Classification & \xmark & \xmark & \xmark & Static unidirectional & Matrix multiplication \\
Mask Classification      & \xmark & \cmark & Cross-attention & Dynamic but asymmetric & Transformer decoder \\
\rowcolor[rgb]{ .925,  .925,  .925}
Offset Learning          & \cmark & \cmark & Dual-decoupled offsets & Elastic bidirectional & Matrix multiplication \\ \bottomrule
\end{tabular}
\label{tab:formulation}
\end{table*}

Semantic segmentation, which aims to assign category labels to every image pixel, plays a vital role in computer vision applications~\cite{cordts2016cityscapes, ronneberger2015u, zhou2018unet++,fan2023advances,lin2023sequential,li2024unbiased,li2024cascadeclip,ji2024segment,yin2024dformer,yin2025dformerv2}.
While recent advances in standard models~\cite{liu2021swin, liu2022swin, zhang2022segvit, strudel2021segmenter, zheng2021rethinking, sung2024contextrast, zhou2024cross} have achieved remarkable segmentation accuracy, their computational and parametric complexity renders them impractical for resource-constrained scenarios.
This has spurred significant interest in efficient semantic segmentation models~\cite{xie2021segformer, guo2022segnext, ni2024context, wan2023seaformer, shim2023feedformer, yu2018bisenet}, which prioritize real-time inference and minimal parameters.

\begin{figure}[!t]
  \centering
  \setlength{\abovecaptionskip}{0pt}
  \includegraphics[width=\columnwidth]{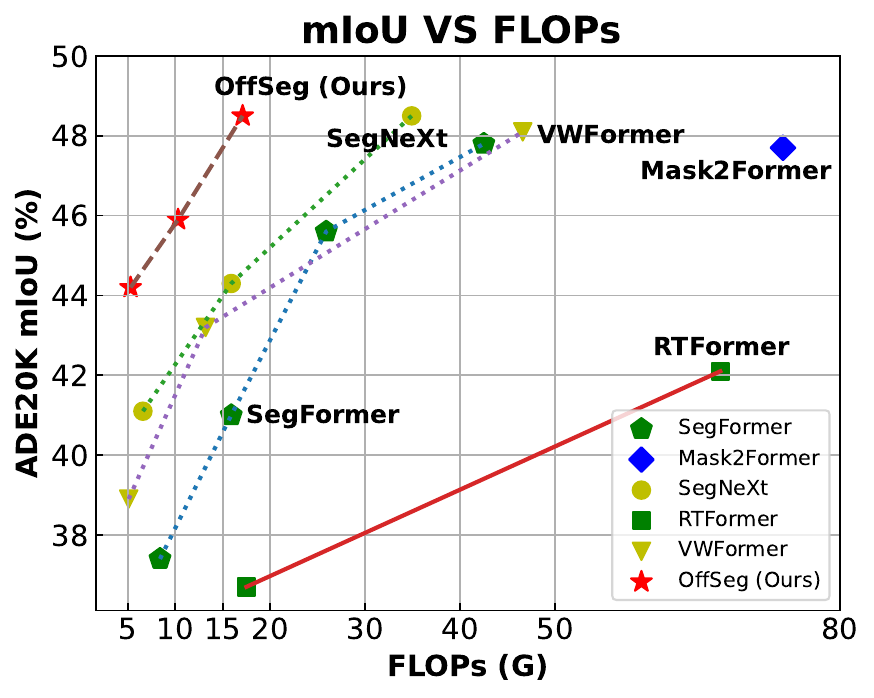}
  \caption{
    Comparisons with popular efficient segmentation methods on the ADE20K 
    \cite{zhou2017scene} dataset. We can see from the figure that our method achieves the best trade-off between performance and computations.
  }\label{fig:cmp_ade20k}
\end{figure}

\begin{figure}[!t]
  \centering
  \small
  \setlength{\abovecaptionskip}{2pt}
  \includegraphics[width=1.\linewidth]{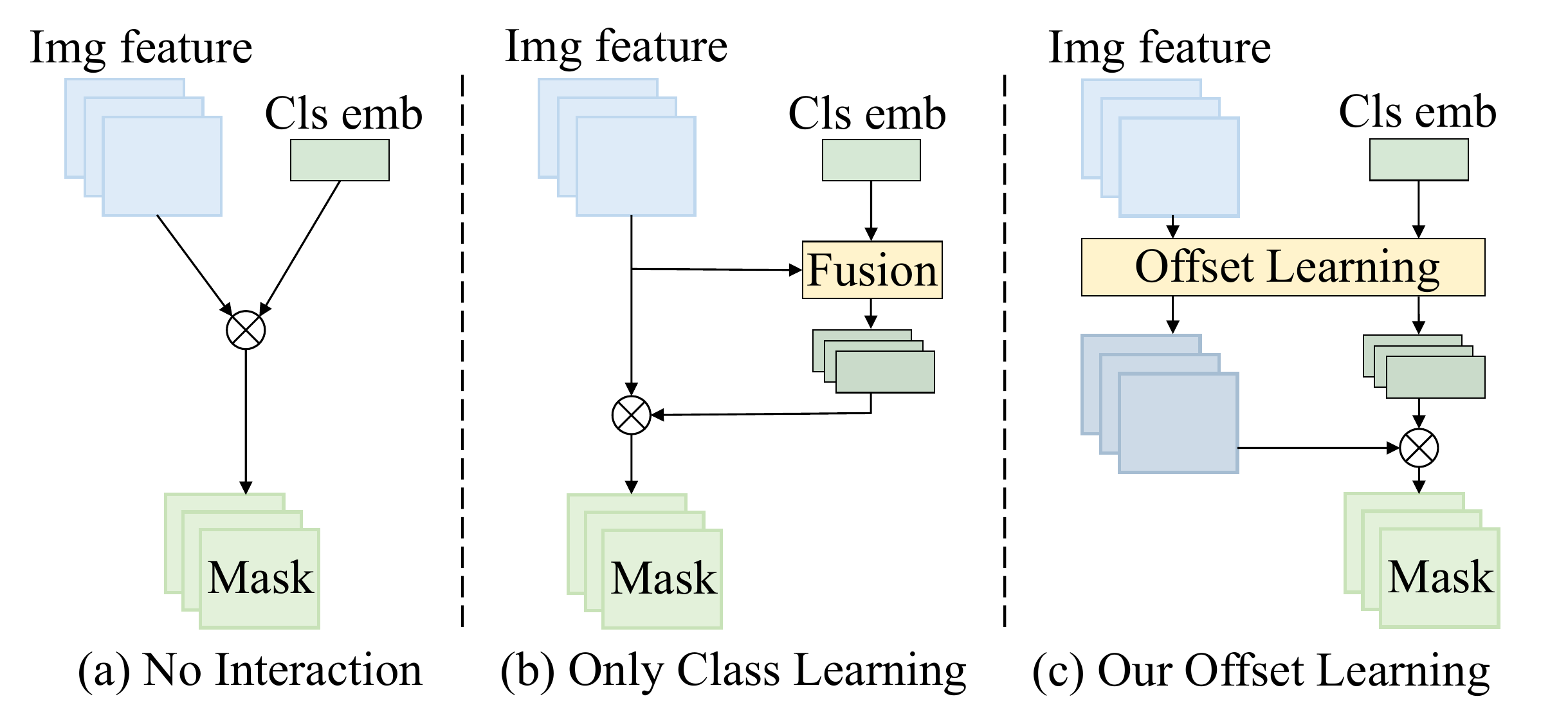}
  \caption{Visual comparison of different semantic segmentation paradigms.
  From left to right, the figures sequentially illustrate per-pixel classification, mask classification, and our proposed offset learning paradigm.}
  \vspace{-5pt}
  \label{fig:compare}
\end{figure}

Conventional segmentation frameworks typically use high-dimensional image features, rich class representations, 
and a large number of parameters, which together yield superior performance compared to more compact architectures.
This phenomenon aligns with the neural scaling law, where model capacity positively correlates with segmentation accuracy until it is able to reach computational resource constraints.
In contrast, efficiency-oriented architectures~\cite{yu2018bisenet, yu2021bisenet, xie2021segformer, wang2022rtformer} face an inherent trade-off: aggressive model compression weakens their capacity to align category semantics with localized visual cues.
This misalignment results in blurred object boundaries, missed small instances, and inconsistent predictions, all of which are further intensified by the prevailing per-pixel classification paradigm (see \figref{fig:compare}(a)).
While existing works employ lightweight backbones~\cite{guo2022segnext, li2022efficientformer, wan2023seaformer} or spatial downsampling~\cite{li2019dfanet, zhang2022topformer, wu2025low} to achieve efficient performance, they largely overlook the fundamental challenge of jointly refining category and feature representations under strict parametric constraints.

To uncover the fundamental issues inherent in the per-pixel classification paradigm, we employ the ideal class representation (feature) mining method to derive optimal class-specific representations for individual images. 
Statistical analysis (\figref{fig:heat}) reveals that the similarity between optimal class representations of the same category across different images is remarkably low. 
This finding shows that using fixed class representations for all images, as in the per-pixel classification approach, is suboptimal since it fails to adapt to the unique image features and class-specific details in each image.

Based on the observations of the fundamental challenge, we propose an offset learning paradigm, 
a novel segmentation method that can explicitly learn and rectify the deviation between class representations and image features through learnable feature offsets (FOs) and class offsets (COs).
Our key insight is that, while efficient models lack sufficient parameters to model ideal category-feature relationships, they can effectively learn to predict the offset between initially coarse representations and their optimal counterparts.
Specifically, our offset learning paradigm consists of two primary branches: the Class Offset Learning branch and the Feature Offset Learning branch. 
These two branches are designed to learn COs and FOs, respectively, enabling flexibility of both image features and class representations.

As shown in \figref{fig:compare}(b), in addition to the per-pixel segmentation paradigm, there exists a mask-based segmentation paradigm~\cite{cheng2021per, cheng2022masked, cavagnero2024pem}, which employs cross-attention to facilitate interaction between learnable queries and image features. 
This approach enables queries to learn image-specific characteristics adaptively. 
However, it has two inherent limitations: (1) It only adjusts the queries while leaving the image features static, and (2) the cross-attention introduces significant computational overhead. 
As summarized in \tabref{tab:formulation}, our method distinguishes itself from these two paradigms through two key advantages: (1) the dual adaptability of both image features and class representations, and (2) negligible interaction overhead.

Based on our proposed offset learning paradigm (\figref{fig:compare}(c)), we design a straightforward segmentation network named~\ours{}, which consists solely of a backbone and a pixel decoder. 
As a plug-and-play paradigm, we apply our framework to SegNeXt~\cite{guo2022segnext} (CNN-based), SegFormer~\cite{xie2021segformer} (Transformer-based), and Mask2Former~\cite{cheng2022masked} (mask classification) to demonstrate its effectiveness and flexibility. 
Extensive experiments across four benchmark datasets show that the results consistently validate the efficiency and effectiveness of our method. 
Performance improvements achieved across different architectures and datasets highlight the robustness and generalizability of our approach.
In \figref{fig:cmp_ade20k}, we present the performance of our model across different scales. 
The results demonstrate that our~\ours{} achieves a superior balance between performance and computational efficiency.

To sum up, our main contributions can be summarized as follows:
\setlength{\parskip}{0pt}
\begin{itemize} [topsep=5px]
  \item We identify the core limitation of per-pixel segmentation through statistical analysis and ideal class representation (feature) mining, exposing the intrinsic misalignment between static image features and class representations.
  \item We propose a parameter-efficient offset learning paradigm with dual branches that jointly adapt image features and class representations with nearly negligible computational overhead. 
  \item Extensive experiments demonstrate superior performance of our proposed \ours{} and effectiveness over previous per-pixel classification (CNN, Transformer) and mask classification paradigms.
\end{itemize}

\section{Related Work}
\label{sec:related}

\subsection{Traditional Semantic Segmentation}
Semantic segmentation has witnessed significant advancements through large-scale models that prioritize accuracy over computational efficiency. 
Pioneering works like fully convolutional networks (FCN)~\cite{long2015fully} established the foundation by replacing fully connected layers with convolutional operations, 
enabling dense pixel-wise predictions. 
With this paradigm established, subsequent CNN-based works~\cite{badrinarayanan2017segnet,ronneberger2015u,yu2015multi,zhao2017pyramid,fu2019dual,yuan2020object,wang2020deep,zhang2023augmented,liang2024relative} have enhanced FCN from various perspectives.
For example, U-Net~\cite{ronneberger2015u} further enhances feature localization through symmetric encoder-decoder structures and skip connections. 
From the perspective of context aggregation,
DeepLab series ~\cite{chen2014semantic, chen2017deeplab, chen2017rethinking, chen2018encoder} employ atrous spatial pyramid pooling (ASPP) to capture multi-scale contextual information.
PSPNet~\cite{zhao2017pyramid} proposes pyramid pooling modules to aggregate global context across different sub-regions. 
Benefiting from the success of attention mechanisms~\cite{vaswani2017attention,dosovitskiy2020image}, Transformer-based approaches~\cite{zheng2021rethinking,yuan2021hrformer,strudel2021segmenter,ranftl2021vision,lee2022mpvit} have achieved remarkable results.
For instance, SERE~\cite{zheng2021rethinking} redefines semantic segmentation as a sequence-to-sequence prediction task, leveraging global self-attention to model full-image context.
Unlike the per-pixel classification paradigm, MaskFormer series~\cite{cheng2021per, cheng2022masked} introduces a mask classification paradigm, where learnable queries interact with image features through a transformer decoder.

\subsection{Efficient Semantic Segmentation}
While traditional models achieve high segmentation accuracy, their computational demands hinder real-time applications, 
driving the development of efficient semantic segmentation works~\cite{xie2021segformer,guo2022segnext,cavagnero2024pem,yu2018bisenet,fan2021rethinking,zhang2022topformer,wu2023p2t,wan2023seaformer,shim2023feedformer,yu2021bisenet,wang2022rtformer, wu2025low}. 
From the perspective of the backbone network, SegFormer~\cite{xie2021segformer} proposes a lightweight and hierarchically structured transformer encoder, while SegNeXt~\cite{guo2022segnext} proposes a more effective convolutional attention solely through multi-scale convolutions to construct an efficient backbone.
LRFormer~\cite{wu2025low} introduces a highly-efficient transformer with linear attention, which is computed in a very low resolution space.
For the decoder, FeedFormer~\cite{shim2023feedformer} employs a transformer that treats image features as queries to extract structural information.
VWFormer~\cite{yan2024multi} augments multi-scale representations by interacting with multiple windows of different scales through cross-attention.
CGRSeg~\cite{ni2024context} utilizes pyramid context-guided spatial feature reconstruction to enhance the ability of foreground objects representation from both horizontal and vertical dimensions.

For efficient segmentation models, since the mask classification paradigm requires a computationally heavy transformer decoder for feature interaction, they all adopt the per-pixel classification segmentation paradigm.
Although per-pixel classification incurs little computational overhead, it inherently suffers from the misalignment between image features and class representations.
The dilemma becomes more pronounced in lightweight scenarios (\figref{fig:heat}), which motivates us to develop a segmentation paradigm tailored for efficient semantic segmentation.

\section{Method}
\label{sec:method}

\newcommand{\AddImg}[1]{\includegraphics[width=.325\linewidth]{figures/#1.pdf}}
\begin{figure}[!t]
  \centering
  \AddImg{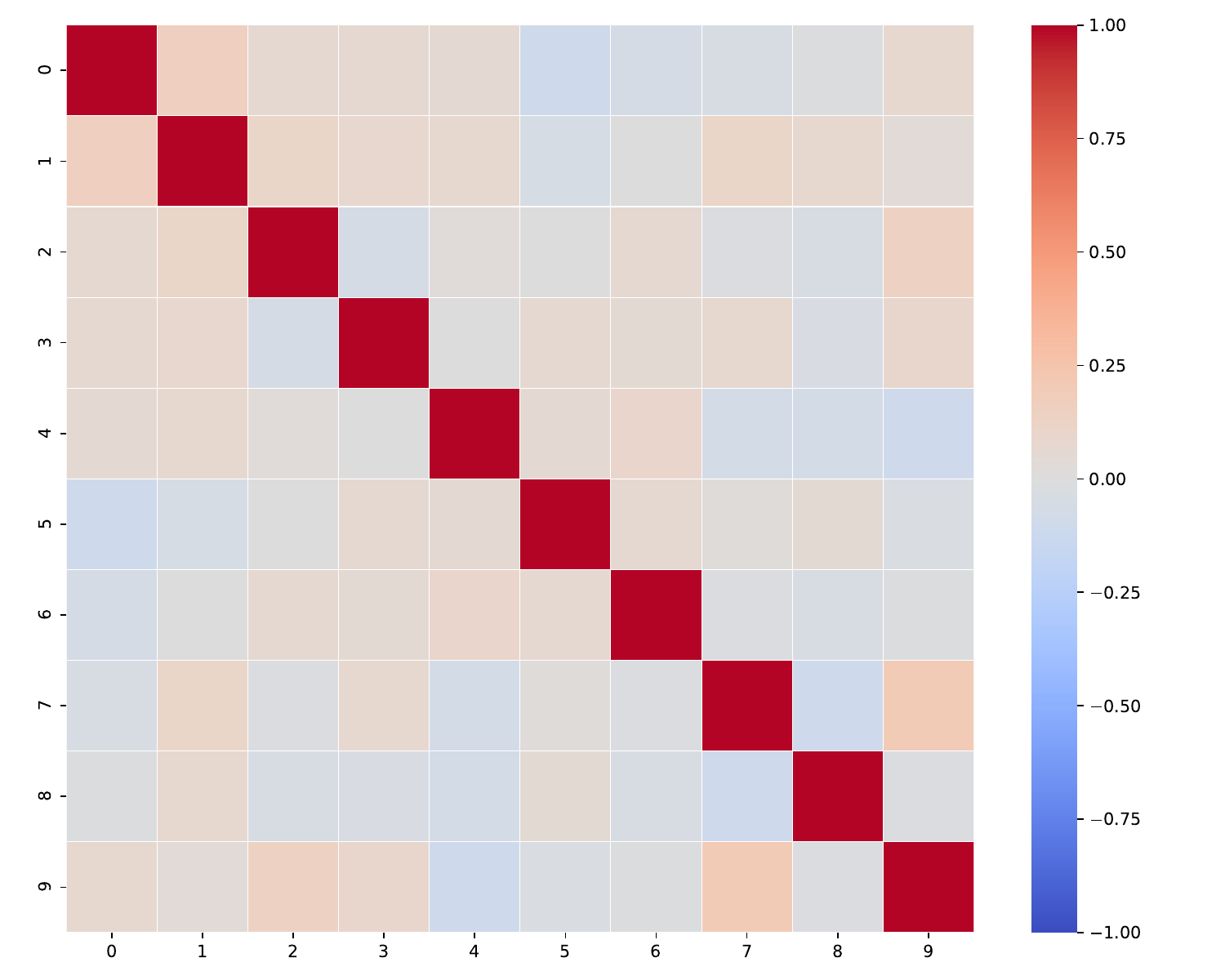} \hfill \AddImg{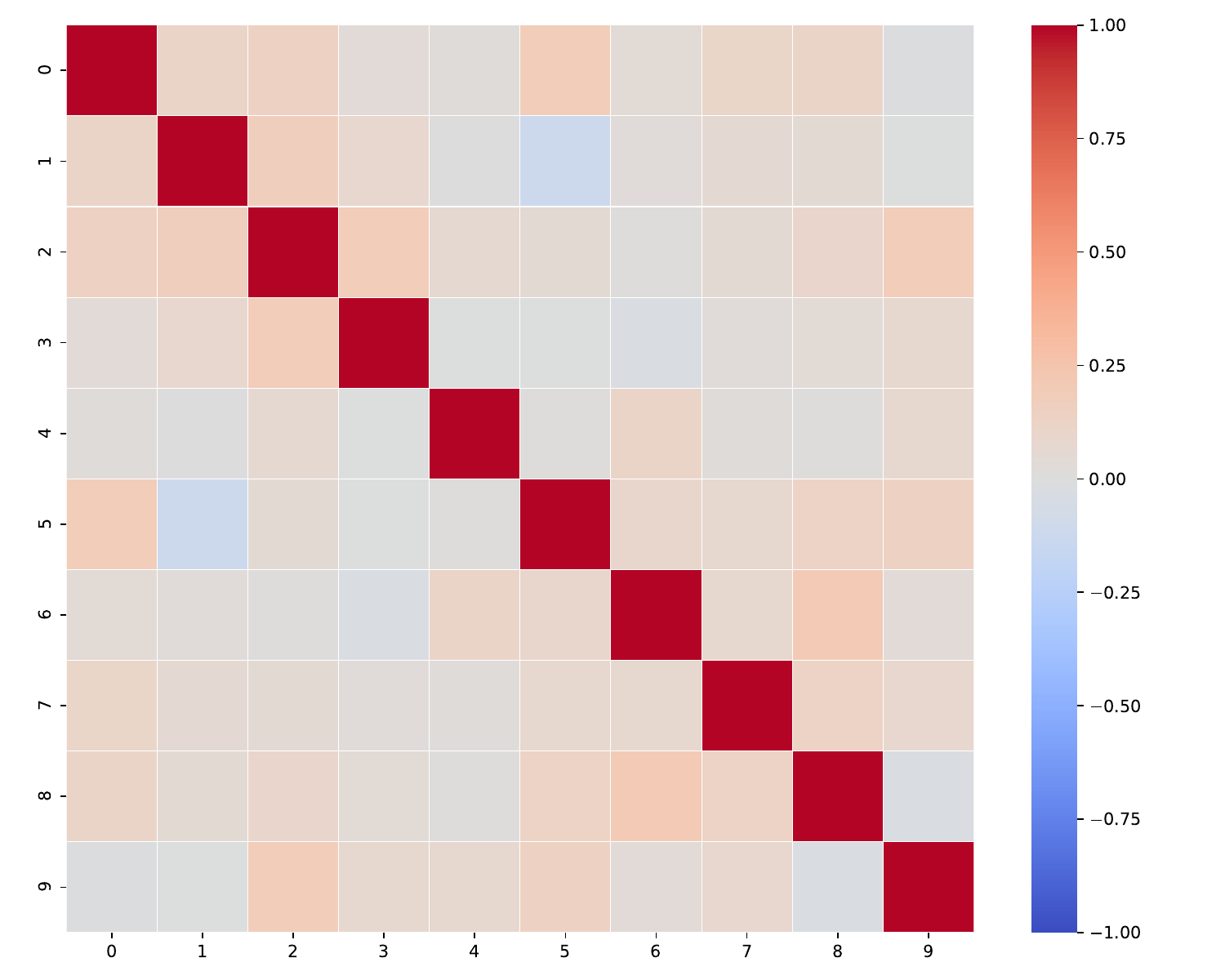} \hfill \AddImg{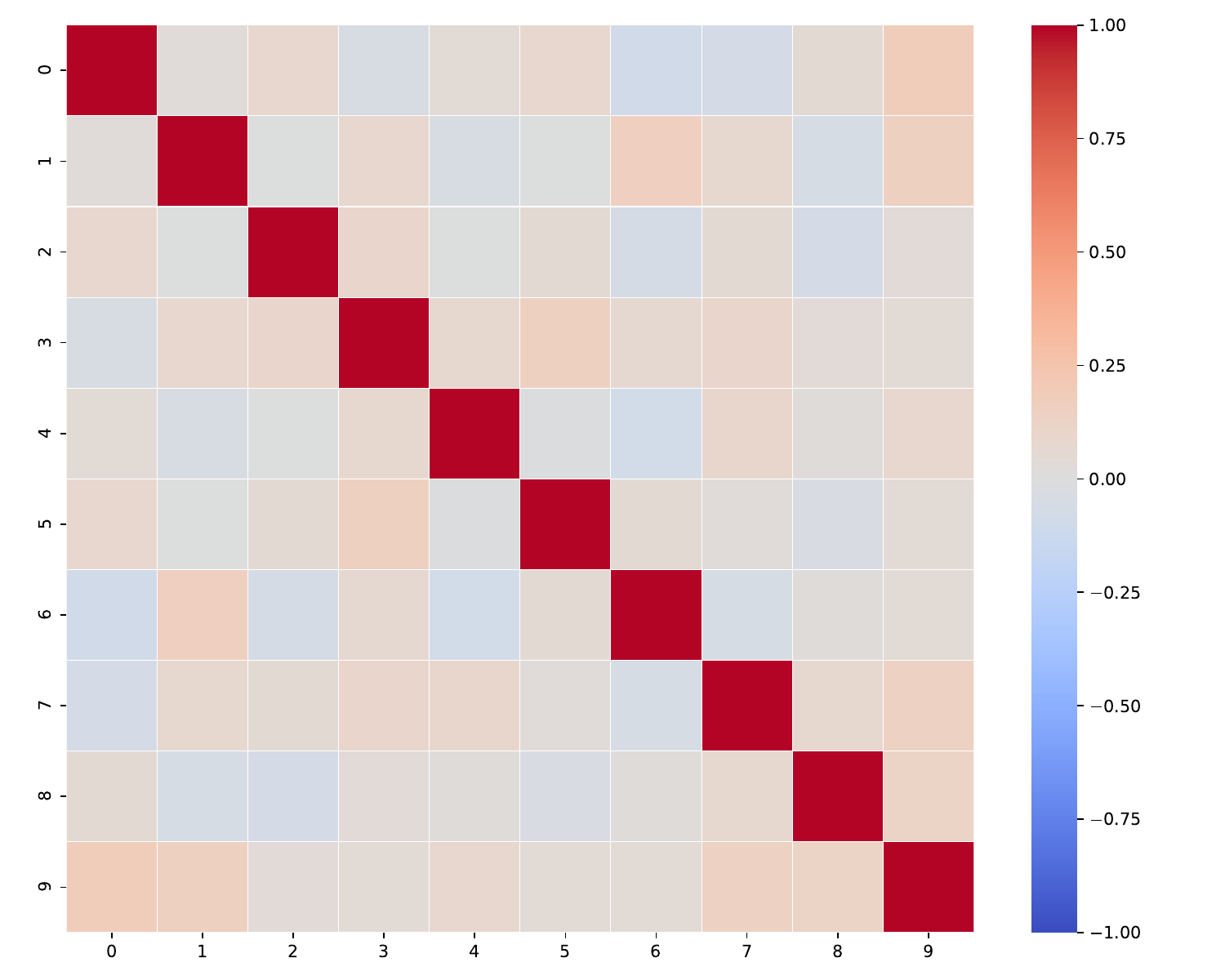} \\ \vspace{-0.06in}
  \leftline{\footnotesize \hspace{0.25in} building \hspace{0.8in} sky \hspace{0.9in} floor }
  \AddImg{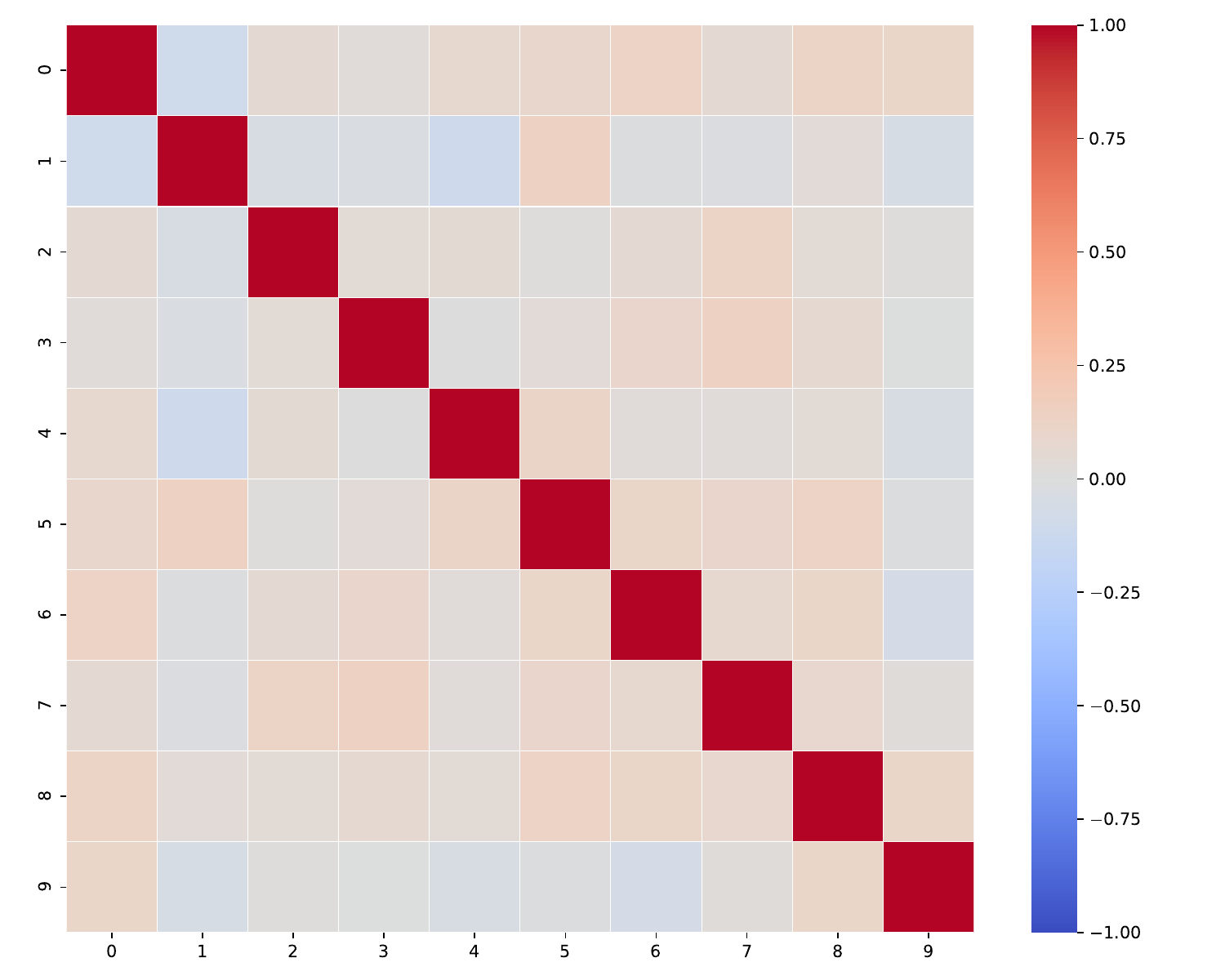} \hfill \AddImg{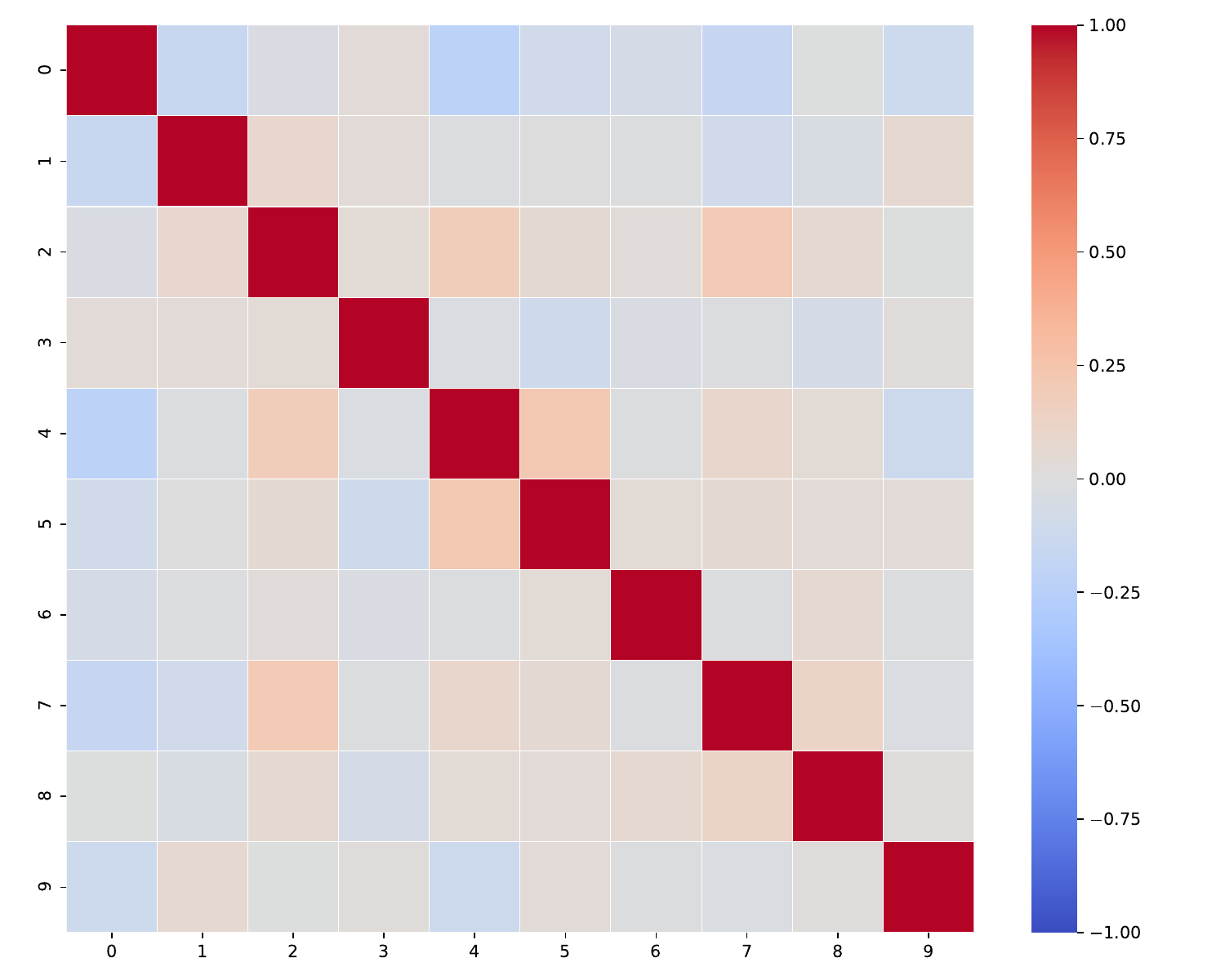} \hfill \AddImg{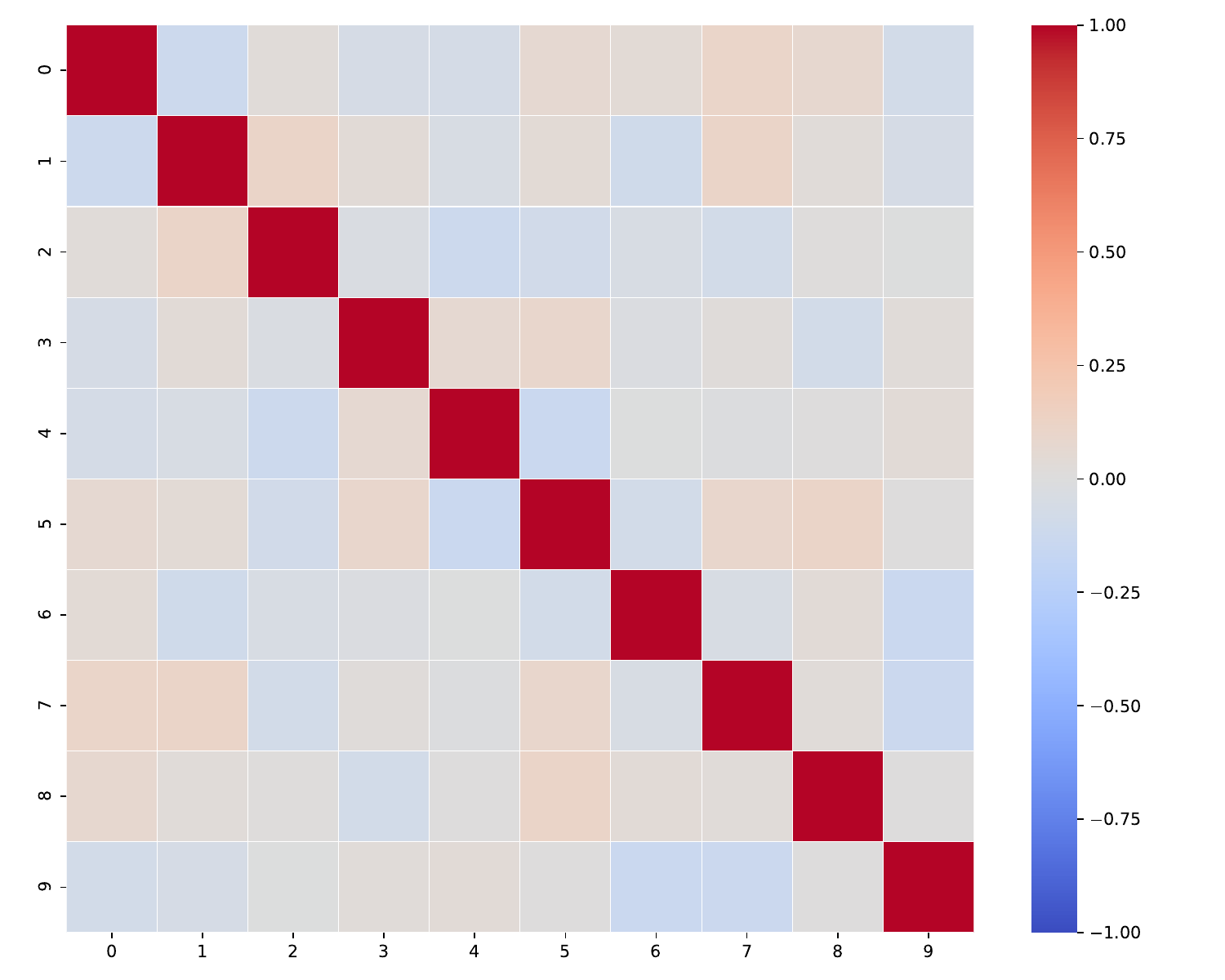} \\ \vspace{-0.06in}
  \leftline{\footnotesize \hspace{0.4in} tree \hspace{0.8in} person \hspace{0.8in} glass }
  \caption{Heatmap visualizations of the ideal class representations similarity.
   We can observe that the correlations between different ideal class representations of the same category are very low.
  }
  \label{fig:heat}
  \vspace{-6pt}
\end{figure}

\begin{figure*}[!t]
  \centering
  \setlength{\abovecaptionskip}{2pt}
  \includegraphics[width=\linewidth]{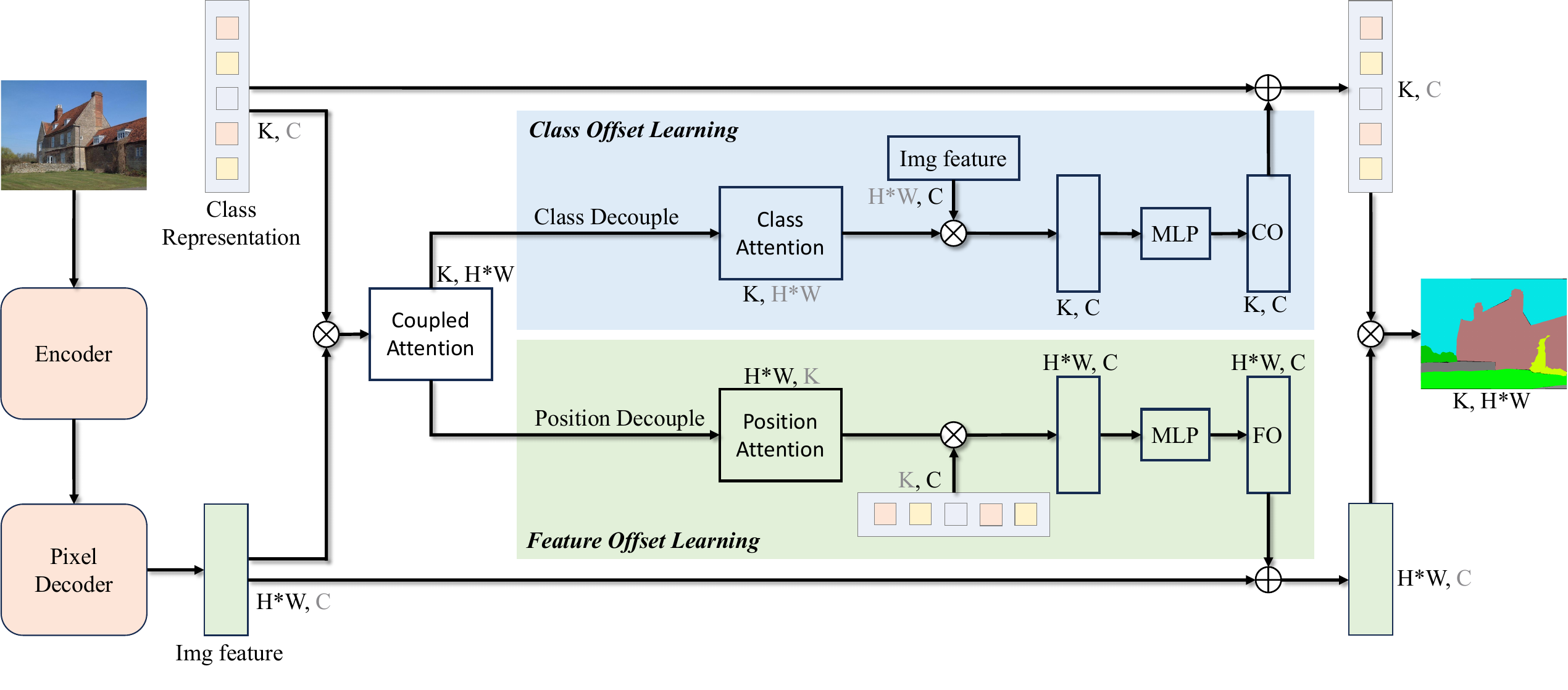}
  \caption{Framework of the proposed~\ours.
  Given an input image, we first use the encoder to extract the multi-scale features and then use the pixel decoder to generate the image feature. The offset learning paradigm contains two branches: Class offset learning branch and Feature offset learning branch. With the learned class offset (CO) and feature offset (FO), we guide the initial feature to aligned space. We denote the dimension of matrix multiplication by grey.
}
  \label{fig:pipeline}
\end{figure*}

\subsection{Revisiting Per-pixel Classification}
Per-pixel classification, the cornerstone of conventional semantic segmentation, independently assigns labels to each pixel by comparing its feature vector with predefined category prototypes.
Traditional per-pixel classification maps pixel embeddings $\mathbf{E} \in \mathbb{R}^{HW \times C}$ to class scores $\mathbf{P}$ via a $1 \times 1$ convolution:
\begin{equation}
\label{equ:conv1x1}
\mathbf{P}_{i,j}=\mathbf{W}_c\cdot\mathbf{E^\top}_{i,j},
\end{equation}
where $\mathbf{W} \in \mathbb{R}^{K \times C}$ is learnable parameter for $K$ classes, $c$ is the class index. This paradigm treats each pixel independently, ignoring contextual correlations.

While widely adopted, this paradigm suffers from two critical issues in efficient segmentation scenarios.
First, the per-pixel classification paradigm relies on fixed class representations to categorize pixels. 
Second, this approach assumes that the network can learn identical features for the same category across different images. 
However, our subsequent analysis in \secref{sec:ideal} demonstrates that this assumption is fundamentally unattainable. 
This misalignment between fixed class representations and diverse image features serves as compelling evidence for the necessity of adaptive mechanisms in modern segmentation frameworks.

\subsection{Ideal Class Representation (Feature) Mining}
\label{sec:ideal}
To theoretically demonstrate the necessity of adaptive class representations and image features, 
we derive optimal per-image class prototypes through inverse reasoning based on ground-truth masks.
Given an input image with ground-truth mask $\mathbf{M} \in \mathbb{R}^{K \times HW}$ and its deep feature $\mathbf{E} \in \mathbb{R}^{HW \times C}$, the ideal class prototypes $\mathbf{W^*} \in \mathbb{R}^{K \times C}$ should satisfy: $\mathbf{M} = \mathbf{W^*} \cdot \mathbf{E^\top}$,
where each row of $\mathbf{W^*}$ represents the optimal prototype for a specific class.
Solving this linear system yields:
\begin{equation}
\label{equ:inverse}
\mathbf{W^*} = \mathbf{M} \cdot \left(\mathbf{E^\top}\right)^{\dagger},
\end{equation}
where $\left(\cdot\right)^{\dagger}$ denotes the Moore-Penrose pseudoinverse.

By recomputing masks via $\mathbf{M}_{\mathrm{pred}}=\mathbf{W}^*\cdot\mathbf{E}^\top$, we achieve near-perfect reconstruction with around 95\% mIoU on the ADE20K dataset, confirming the theoretical validity of $\mathbf{W^*}$.
With such a mathematical derivation, we conduct a similarity analysis using SegFormer~\cite{xie2021segformer} (an efficient network with 4.3M learnable parameters).
Firstly, we randomly select six categories (\ie, building, sky, floor, tree, person, glass) and compute $\mathbf{W^*}$ for 10 images per category for a better view.
As shown in \figref{fig:heat}, we visualize their pairwise similarity via heatmaps.
Strikingly, the ideal class representations exhibit less similarity than our common sense.
The low correlation heatmap patterns reveal that optimal class representations vary drastically across images for the same class.
This phenomenon stems from a fundamental tension in efficient models: aggressive feature compression amplifies intra-class feature variance, forcing $\mathbf{W^*}$ to diverge significantly to fit distorted features.
The above analysis reveals two critical implications: 
\begin{itemize}
    \item Fixed prototypes fail: Fixed class representations cannot universally align with highly variable features of different images in efficient models.
    \item Fixed features fail: The relationship $\mathbf{W}^*\propto f(\mathbf{E})$ implies that distorted features also hinder prototype stability, which needs a vicious cycle requiring joint correction.
\end{itemize}

\subsection{Offset Learning Paradigm}

Our offset learning paradigm redefines the segmentation paradigm as a dual-decoupled alignment process:
\begin{equation}
\label{equ:formulation}
\mathbf{M}=(\mathbf{W}+\Delta\mathbf{W})\cdot(\mathbf{E}+\Delta\mathbf{E})^\top,
\end{equation}
which departs our method from conventional per-pixel classification~\cite{chen2017deeplab,guo2022segnext,hou2020strip,zhao2017pyramid} and mask-centric approaches~\cite{cheng2021per,cheng2022masked}. 
The core innovation lies in the class offset learning and feature offset learning branches that collaboratively refine class prototypes and spatial features through decoupled attention mechanisms as shown in \figref{fig:pipeline}.

To be specific, given the image feature $\mathbf{E}\in\mathbb{R}^{HW\times C}$ and class embedding $\mathbf{W}\in\mathbb{R}^{K\times C}$,
we compute a coupled attention matrix $\mathbf{A}_c$:
\begin{equation}
\label{equ:coupled_attention}
\mathbf{A}_c=\mathbf{W}\cdot\mathbf{E}^\top,
\end{equation}
where $\mathbf{A}_c\in\mathbb{R}^{K\times HW}$.
This matrix encodes correlations between classes and spatial positions and is used in the subsequent class offset learning branch and feature offset learning branch.

\myPara{Class offset learning} dynamically adjusts class representations based on spatial context, alleviating the rigidity of fixed class embeddings.
First, we apply the softmax normalization along spatial dimensions $\mathrm{Softmax}_{S}$ to generate class-wise attention weights:
\begin{equation}
\label{equ:class_decouple}
\mathbf{A}_{\mathrm{cls}}=\mathrm{Softmax}_{S}(\mathbf{A}_{c})\in\mathbb{R}^{K\times HW},
\end{equation}
where each row $\mathbf{a}_{k}\in\mathbf{A}_{\mathrm{cls}}$ indicates the spatial importance distribution for class $k$.
Then, we aggregate spatial features weighted by class attention:
\begin{equation}
\mathbf{F}_{\mathrm{cls}}=\mathbf{A}_{\mathrm{cls}}\cdot\mathbf{E}\in\mathbb{R}^{K\times C},
\end{equation}
where $\mathbf{F}_{\mathrm{cls}}$ contains class-specific prototypes that encode global spatial distributions.
Finally, we generate class offsets via an MLP: 
\begin{equation}
\Delta\mathbf{W}=\mathrm{MLP}(\mathbf{F}_{\mathrm{cls}})\in\mathbb{R}^{K\times C},
\end{equation}
and adjust original representations as follows:
\begin{equation}
\mathbf{W}_{\mathrm{adj}}=\mathbf{W}+\Delta\mathbf{W}.
\end{equation}
This branch learns image-specific offsets to align class embeddings with corresponding image features, narrowing their representational gap.

\myPara{Feature offset learning} refines image features by injecting class-aware semantics. The intention is to overcome the local ambiguity in per-pixel classification.
As shown in \figref{fig:pipeline}, the feature offset learning branch is dual to the class offset learning branch.

To be specific, like the class offset learning branch, we first apply softmax along the class dimension $\mathrm{Softmax}_K$ and transpose for spatial alignment:
\begin{equation}
\mathbf{A}_{\mathrm{pos}}=\left(\mathrm{Softmax}_K(\mathbf{A}_c)\right)^T\in\mathbb{R}^{HW\times K},
\end{equation}
where each row $\mathbf{a}_{i}\in\mathbf{A}_{\mathrm{pos}}$ represents the class probability distribution at position $i$.
Then, we fuse class semantics into spatial positions via the following equation:
\begin{equation}
\mathbf{F}_{\mathrm{pos}}=\mathbf{A}_{\mathrm{pos}}\cdot\mathbf{W}\in\mathbb{R}^{HW\times C},
\end{equation}
where $\mathbf{F}_{\mathrm{pos}}$ encodes position-wise semantic guidance from all classes.
Finally, we adopt an MLP to generate feature offset:
\begin{equation}
\Delta\mathbf{E}=\mathrm{MLP}(\mathbf{F}_{\mathrm{pos}})\in\mathbb{R}^{HW\times C},
\end{equation}
and use it to guide the original features:
\begin{equation}
\mathbf{E}_{\mathrm{adj}}=\mathbf{E}+\Delta\mathbf{E}.
\end{equation}
The final segmentation masks are generated through bidirectional elastic alignment:
\begin{equation}
\mathbf{M}=\mathbf{W}_{\mathrm{adj}}\cdot\mathbf{E}_{\mathrm{adj}}^T,
\end{equation}
which is another form of \eqnref{equ:formulation}.

We have summarized the main differences between our offset learning paradigm and other semantic segmentation paradigms in~\tabref{tab:formulation}.
Different from per-pixel classification (\eg, SegFormer~\cite{xie2021segformer}, SegNeXt~\cite{guo2022segnext}), 
which relies on static alignment between fixed features and rigid class embeddings,
or mask classification (\eg, MaskFormer~\cite{cheng2021per}, Mask2Former~\cite{cheng2022masked}) that dynamically refines class queries by a heavy transformer decoder, our framework uniquely introduces bidirectional offset learning with even negligible learnable parameters.
This method enables symmetric adaptation: class representations can be adjusted through class-specific spatial prototypes, while image features can be refined by position-aware semantic guidance.
By decoupling class- and position-wise interactions into two distinct pathways, our method achieves elastic feature-class alignment, where both modalities co-evolve to capture instance-specific geometries and contextual semantics.
This contrasts sharply with the unidirectional or hard-coded alignment strategies in existing paradigms.

\subsection{Overall Architecture}
To validate the efficiency and effectiveness of our proposed offset learning paradigm, we design a standard semantic segmentation model with the following efficient components without structural modifications.
For the backbone, we employ a hybrid architecture named EfficientFormerV2~\cite{li2023rethinking}, which achieves a balance between parameter efficiency and performance through a fine-grained joint search strategy.
For multi-scale feature aggregation, we select FreqFusion~\cite{chen2024frequency}, which fuses two scale features with frequency-aware operators.
Notably, when combined with our offset learning paradigm, the entire model introduces nearly negligible learnable parameters (0.1-0.2M).

\section{Experiments}
\label{sec:experiments}

\subsection{Experimental Settings}

\begin{table*}
\centering
\setlength{\abovecaptionskip}{2pt}
\small
\caption{Performance comparison of state-of-the-art methods on ADE20K, Cityscapes and COCO-Stuff datasets. FLOPs (G) is computed at input resolutions of 512$\times$512 for ADE20K and COCO-Stuff, and 2048$\times$1024 for Cityscapes.}
\begin{tabular}{l|c|cc|cc|cc}
\toprule
\multirow{2}{*}{Method} & \multirow{2}{*}{Params (M)} & \multicolumn{2}{c|}{ADE20K} & \multicolumn{2}{c|}{Cityscapes} & \multicolumn{2}{c}{COCO-Stuff} \\
                 &             & FLOPs (G) & mIoU & FLOPs (G) & mIoU & FLOPs (G) & mIoU \\ \midrule
SegFormer-B0~\cite{xie2021segformer}     & 3.8 & 8.4 & 37.4 & 125.5 & 76.2 & 8.4 & 35.6 \\
RTFormer-Slim~\cite{wang2022rtformer}    & 4.8 & 17.5 & 36.7 & -  & 76.3 & - & - \\
FeedFormer-B0~\cite{shim2023feedformer}  & 4.5 & 7.8 & 39.2 & 107.4 & 77.9 & -   & -    \\
Seaformer-L~\cite{wan2023seaformer}      & 14.0 & 6.5 & 42.7 & -     & -    &  -  & -    \\
VWFormer-B0~\cite{yan2024multi}          & 3.7 & 5.1 & 38.9 & -     & 77.2 & 5.1 & 36.2 \\
CGRSeg-T~\cite{ni2024context}            & 9.4 & 4.0 & 43.6 & -     & -    & 4.0 & 42.2 \\
EDAFormer-T~\cite{yu2024embedding}       & 4.9 & 5.6 & 42.3 & 151.7 & 78.7 & 5.6 & 40.3 \\
\rowcolor[rgb]{ .925,  .925,  .925}
\ours-T                                  & 6.2 & 5.3 & 44.2 & 44.8  & 78.9 & 5.3 & 41.9 \\ \midrule
SegFormer-B1~\cite{xie2021segformer}     & 13.7 & 15.9 & 42.2 & 243.7 & 78.5 & 15.9 & 40.2 \\
SegNeXt-S~\cite{guo2022segnext}          & 13.9 & 15.9 & 44.3 & 124.6  & 81.3 & 15.9 & 42.2 \\
RTFormer-Base~\cite{wang2022rtformer}    & 16.8 & 67.4 & 42.1 & -  & 79.3 & 26.6 & 35.3 \\
VWFormer-B1~\cite{yan2024multi}          & 13.7 & 13.2 & 43.2 & -     & 79.0 &  -  & 41.5 \\
PEM-STDC1~\cite{cavagnero2024pem}        & 17.0 & 16.0 & 39.6 & -     & -    &  -  & -    \\
\rowcolor[rgb]{ .925,  .925,  .925}
\ours-B                                  & 13.0 & 10.3 & 45.9 & 86.5  & 80.5  & 10.3   & 44.3  \\ \midrule
SenFormer~\cite{bousselham2021efficient} & 59.0 & 179.0 & 46.0 & -      & -    &-     & -    \\
SegFormer-B2~\cite{xie2021segformer}     & 27.5 & 25.9 & 45.6 & 717.1 & 81.0 & 26.0 & 44.6 \\
MaskFormer~\cite{cheng2021per}           & 42.0 & 55.0 & 46.7 & -      & -    &-     & -    \\
Mask2Former~\cite{cheng2022masked}       & 47.0 & 74.0 & 47.7 & -      & -    &-     & -    \\
FeedFormer-B2~\cite{shim2023feedformer}     & 29.1 & 42.7 & 48.0 & 522.7 & 81.5 & - & -    \\
PEM-STDC2~\cite{cavagnero2024pem}        & 21.0 & 19.3 & 45.0 & -     & -    &  -  & -    \\
\rowcolor[rgb]{ .925,  .925,  .925}
\ours-L                                  & 26.4 & 17.1 & 48.5 & 143.4  & 81.6 & 17.1   & 46.0  \\ \bottomrule
\end{tabular}
\label{tab:comparison}
\end{table*}

\myPara{Datesets.}
We evaluate our method on four widely adopted semantic segmentation benchmarks: ADE20K~\cite{zhou2017scene}, Cityscapes~\cite{cordts2016cityscapes}, COCO-Stuff~\cite{caesar2018coco}, and Pascal Context~\cite{mottaghi2014role}.
ADE20K~\cite{zhou2017scene} is a scene parsing dataset with 150 object/stuff categories, 
containing 20K/2K/3K images for training/validation/testing. 
It features diverse indoor and outdoor scenes with complex occlusions.
Cityscapes~\cite{cordts2016cityscapes} focuses on urban driving scenarios, providing 5,000 high-resolution images (2048$\times$1024) with 19 semantic classes.
COCO-Stuff~\cite{caesar2018coco} comprises 118K training and 5K validation images with 171 classes (80 things + 91 stuff). 
Its long-tailed distribution challenges model generalization.
PASCAL Context~\cite{mottaghi2014role} dataset comprises 59 semantic categories as foreground objects, with 4,996 training images and 5,104 validation images.

\myPara{Implementation details.}
Our implementation is based on the MMSegmentation~\cite{mmseg2020} with PyTorch. 
Following previous works~\cite{xie2021segformer, guo2022segnext, ni2024context, cheng2021per, cheng2022masked}, we adopt the AdamW~\cite{loshchilov2017decoupled} optimizer with poly learning rate decay and 1,500 iterations linear warmup for all models, without specific tuning for any other settings. 
The batch size is set to 16 for the ADE20K/COCO-Stuff/Pascal Context datasets and 8 for the Cityscapes dataset. 
During training, the image size is cropped to 512$\times$512 for ADE20K and COCO-Stuff, 480$\times$480 for Pascal Context, and 1024$\times$1024 for the Cityscapes dataset.
We adopt the standard data augmentation and train 160k iterations on ADE20K and Cityscapes datasets and 80k iterations on the COCO-Stuff and Pascal Context datasets. 
During inference, we employ single-scale testing for all datasets. 
All experiments are conducted on 8 NVIDIA RTX 3090 GPUs.

\subsection{Main Results}
We evaluate our method on three standard semantic segmentation benchmarks: ADE20K, Cityscapes and COCO-Stuff, as detailed in~\tabref{tab:comparison}.
For the ADE20K dataset, the proposed~\ours-T attains 44.2 mIoU on ADE20K, surpassing EDAFormer-T by 1.9 mIoU while reducing computations by 24\%. 
\ours-B establishes a strong accuracy-efficiency trade-off: 45.9 mIoU on ADE20K (10.3G FLOPs), outperforming SegNeXt-S (+1.6) and PEM-STDC1 (+6.3) with 35\% lower FLOPs than SegNeXt-S.
At the large scale, \ours-L achieves a 48.5 mIoU score on ADE20K with 17.1G FLOPs, outperforming Mask2Former (+0.8) with 4.3$\times$ fewer FLOPs

On the Cityscapes dataset, our \ours-L achieves superior results while using only a quarter of the computational cost required by FeedFormer-B2.
On the COCO-Stuff dataset, our \ours-B surpasses RTFormer-Base by 9.0 mIoU with less than half of its computational cost.

These experimental results suggest that our dual-decoupled offset learning paradigm can effectively address the misalignment between class representations and image features, particularly in class-dense and challenging scenarios, like ADE20K and COCO-Stuff.

\subsection{Generalization Ability}

\begin{table*}
\centering
\small
\setlength{\abovecaptionskip}{2pt}
\setlength\tabcolsep{4.5pt}
\caption{Performance comparison of SegNeXt~\cite{guo2022segnext} and SegNeXt w/ offset learning on ADE20K, Cityscapes, Pascal Context, and COCO-Stuff datasets. 
FLOPs (G) is computed at input resolutions of 2048$\times$1024 for Cityscapes and 512$\times$512 for other datasets.
}
\begin{tabular}{lc|c|cc|cc|cc|cc}
\toprule
\multirow{2}{*}{Method} & \multirow{2}{*}{Offset} & \multirow{2}{*}{Params (M)} & \multicolumn{2}{c|}{ADE20K} & \multicolumn{2}{c|}{Cityscapes} & \multicolumn{2}{c|}{Pascal Context} & \multicolumn{2}{c}{COCO-Stuff} \\
                 &             & & FLOPs (G) & mIoU & FLOPs (G) & mIoU & FLOPs (G) & mIoU & FLOPs (G) & mIoU \\ \midrule
SegNeXt-T &              & 4.3 & 6.6 & 41.1 & 50.5 & 79.8 & 6.6 & 51.2 & 6.6 & 38.7 \\
\rowcolor[rgb]{ .925,  .925,  .925}
SegNeXt-T & $\checkmark$ & 4.4 & 7.2 & 43.0\textsubscript{(+1.9)} & 53.1 & 80.0\textsubscript{(+0.2)} & 6.8  & 53.2\textsubscript{(+2.0)} & 7.3 & 40.0\textsubscript{(+1.3)} \\ \midrule
SegNeXt-S &              & 13.9 & 15.9 & 44.3 & 124.6 & 81.3 & 15.9 & 54.2 & 15.9 & 42.2 \\
\rowcolor[rgb]{ .925,  .925,  .925}
SegNeXt-S & $\checkmark$ & 14.1 & 16.5 & 45.6\textsubscript{(+1.3)} & 127.2 & 81.7\textsubscript{(+0.4)} & 16.1 & 55.9\textsubscript{(+1.7)} & 16.6 & 43.5\textsubscript{(+1.3)} \\ \midrule
SegNeXt-B &              & 27.6 & 34.9 & 48.5 & 275.7 & 82.6 & 34.9 & 57.0 & 34.9 & 45.8 \\
\rowcolor[rgb]{ .925,  .925,  .925}
SegNeXt-B & $\checkmark$ & 28.2 & 34.8 & 49.4\textsubscript{(+0.9)} & 269.6 & 82.8\textsubscript{(+0.2)} & 34.1 & 58.0\textsubscript{(+1.0)} & 35.0 & 45.8\textsubscript{(+0.0)} \\ \bottomrule
\end{tabular}
\label{tab:segnext}
\end{table*}

\begin{table}
\centering
\small
\setlength{\abovecaptionskip}{2pt}
\setlength\tabcolsep{1.5pt}
\caption{Performance comparison of SegFormer~\cite{xie2021segformer} and SegFormer w/ offset learning on ADE20K and COCO-Stuff datasets. FLOPs (G) is computed at input resolutions of 512$\times$512 for all datasets.}
\begin{tabular}{lc|c|cc|cc}
\toprule
\multirow{2}{*}{Method} & \multirow{2}{*}{Offset} & \multirow{2}{*}{Params} & \multicolumn{2}{c|}{ADE20K}  & \multicolumn{2}{c}{COCO-Stuff} \\
                              &              &     & FLOPs & mIoU                       & FLOPs & mIoU \\ \midrule
SegFormer-B0 &              & 3.8M & 8.4       & 37.4                       & 8.6       & 35.6 \\
\rowcolor[rgb]{ .925,  .925,  .925}
SegFormer-B0 & $\checkmark$ & 3.9M & 8.8      & 40.1\textsubscript{(+2.7)} & 8.9       & 38.3\textsubscript{(+2.7)} \\ \midrule
SegFormer-B1 &              & 13.7M & 15.9     & 41.0                       & 16.1      & 40.2 \\
\rowcolor[rgb]{ .925,  .925,  .925}
SegFormer-B1 & $\checkmark$ & 13.9M & 16.3     & 43.7\textsubscript{(+2.7)} & 16.4      & 41.9\textsubscript{(+1.7)} \\ \midrule
SegFormer-B2 &              & 24.8M & 25.9     & 45.6                       & 26.0      & 44.6 \\
\rowcolor[rgb]{ .925,  .925,  .925}
SegFormer-B2 & $\checkmark$ & 24.9M & 26.1     & 47.3\textsubscript{(+1.7)} & 26.2      & 45.2\textsubscript{(+0.6)} \\ \midrule
SegFormer-B3 &              & 44.6M & 42.5     & 47.8                       & 42.6      & 45.5 \\
\rowcolor[rgb]{ .925,  .925,  .925}
SegFormer-B3 & $\checkmark$ & 44.8M & 42.8     & 49.5\textsubscript{(+1.7)} & 42.9      & 46.3\textsubscript{(+0.8)} \\ \midrule
SegFormer-B4 &              & 61.4M & 59.2     & 48.5                       & 59.3      & 46.5 \\
\rowcolor[rgb]{ .925,  .925,  .925}
SegFormer-B4 & $\checkmark$ & 61.6M & 59.5     & 50.1\textsubscript{(+1.6)} & 59.6      & 47.0\textsubscript{(+0.5)} \\ \midrule
SegFormer-B5 &              & 82.0M & 75.2     & 49.1                       & 75.3      & 46.7 \\
\rowcolor[rgb]{ .925,  .925,  .925}
SegFormer-B5 & $\checkmark$ & 82.2M & 75.5     & 50.6\textsubscript{(+1.5)} & 75.5      & 47.2\textsubscript{(+0.5)} \\ \bottomrule
\end{tabular}
\label{tab:segformer}
\vspace{-5pt}
\end{table}

To validate the broad applicability of our offset learning paradigm, 
we integrate it into three representative models: SegNeXt~\cite{guo2022segnext} (CNN-based), SegFormer~\cite{xie2021segformer} (Transformer-based), and Mask2Former~\cite{cheng2022masked} (mask classification). 
For the per-pixel classification framework models (SegNeXt and SegFormer), we adapt our approach by simply replacing the final 1$\times$1 convolutional layer with our offset learning paradigm. 
For the mask classification framework (Mask2Former), we leverage offset learning to align mask embeddings with per-pixel embeddings while remaining other parts unchanged.

\myPara{SegNeXt with offset learning.} 
To evaluate the robustness of our method, we conduct experiments on four datasets. 
As shown in Table~\ref{tab:segnext}, integrating our paradigm into SegNeXt yields consistent performance improvements with negligible parameter overhead.
In summary, our method achieves average improvements of 1.4, 1.2, and 0.5 mIoU across all datasets for the Tiny, Small, and Base scales, respectively, while introducing only 
0.1-0.2M
additional parameters. 
These results demonstrate the effectiveness and efficiency of our approach in enhancing segmentation performance.
We also evaluate SegNeXt-T with offset learning paradigm using the ensemble strategy (multi-scale).
The model achieves mIoU of 43.2 on ADE20K, 81.9 on Cityscapes, 54.5 on Pascal Context, and 40.5 on COCO-Stuff, further enhancing segmentation accuracy.

To further show the advantages of our method, we show the segmentation results based on the SegNeXt-T model in \figref{fig:vis_segnext}. The visualizations reveal that our model achieves more precise segmentation outcomes, particularly in the identification of background regions and small objects. This qualitatively validates that our method can better align image features with class representations.

Furthermore, the diminishing improvement trend (from 1.4 to 0.5 mIoU) as the model size increases suggests that larger-scale models inherently possess superior feature-representation alignment capabilities. 
This experimental observation directly validates our hypothesis of a positive correlation between model size and class alignment capability.
The results also demonstrate that our approach resolves the identified issues with the efficiency level anticipated in our analysis.

\renewcommand{\AddImg}[1]{\includegraphics[width=.24\linewidth]{figures/vis/#1}}
\begin{figure}[!t]
  \centering
  \AddImg{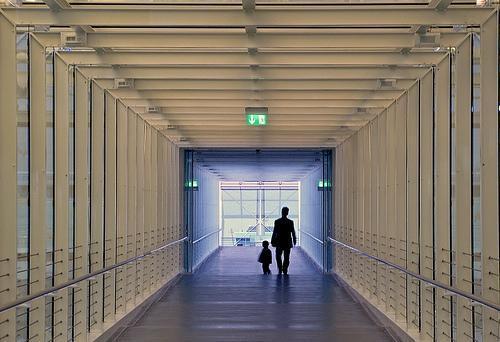} \hfill \AddImg{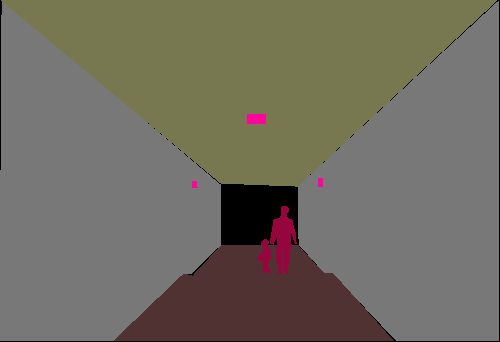} \hfill \AddImg{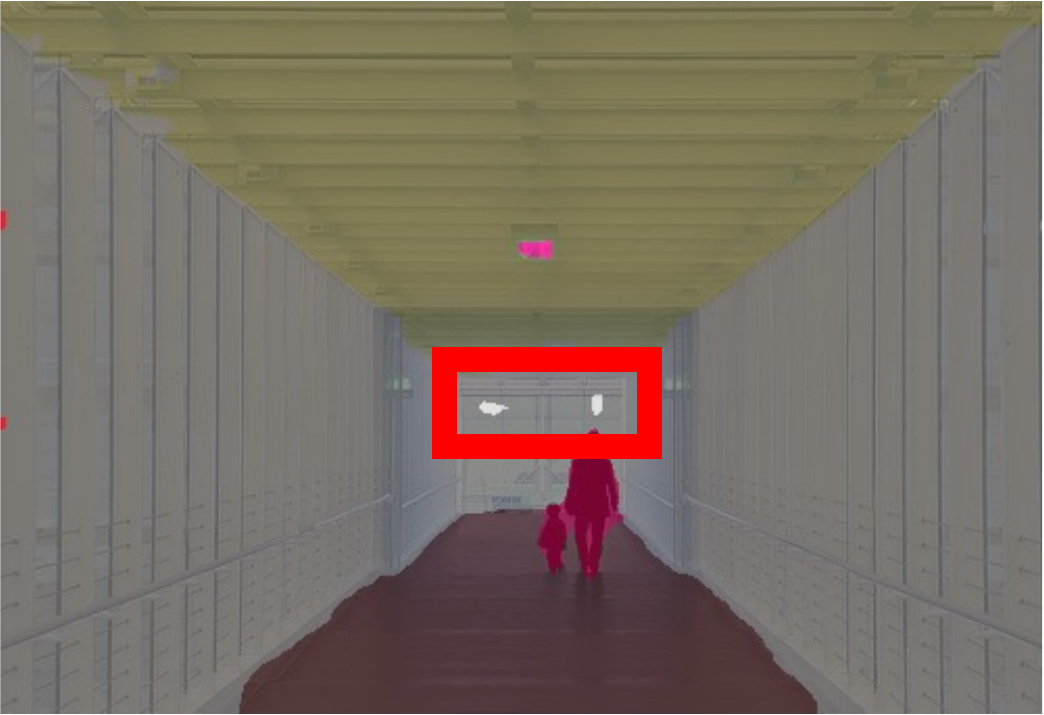} \hfill \AddImg{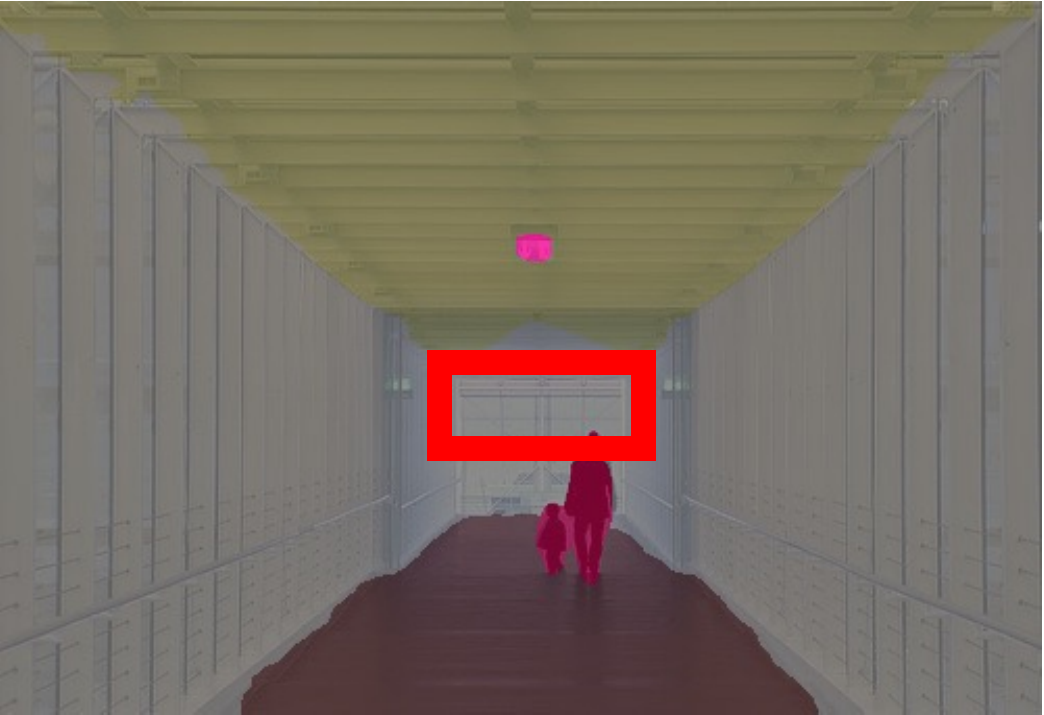} \\ 
  \AddImg{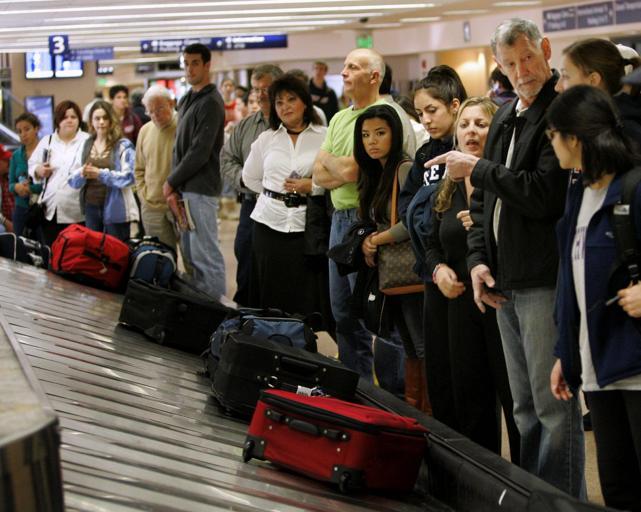} \hfill \AddImg{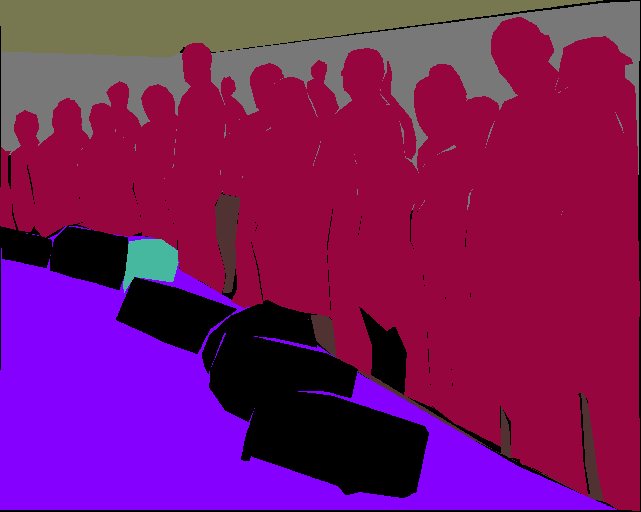} \hfill \AddImg{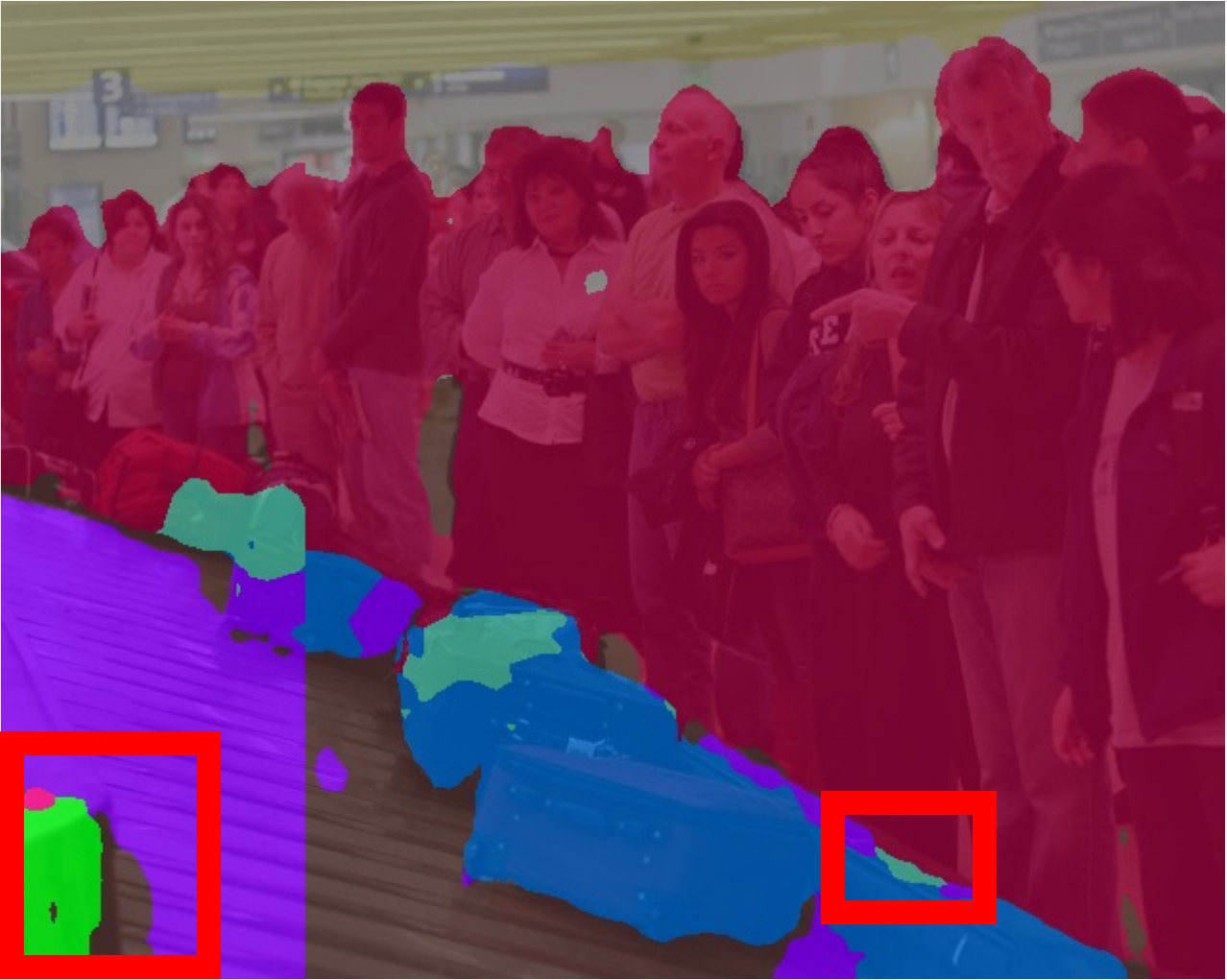} \hfill \AddImg{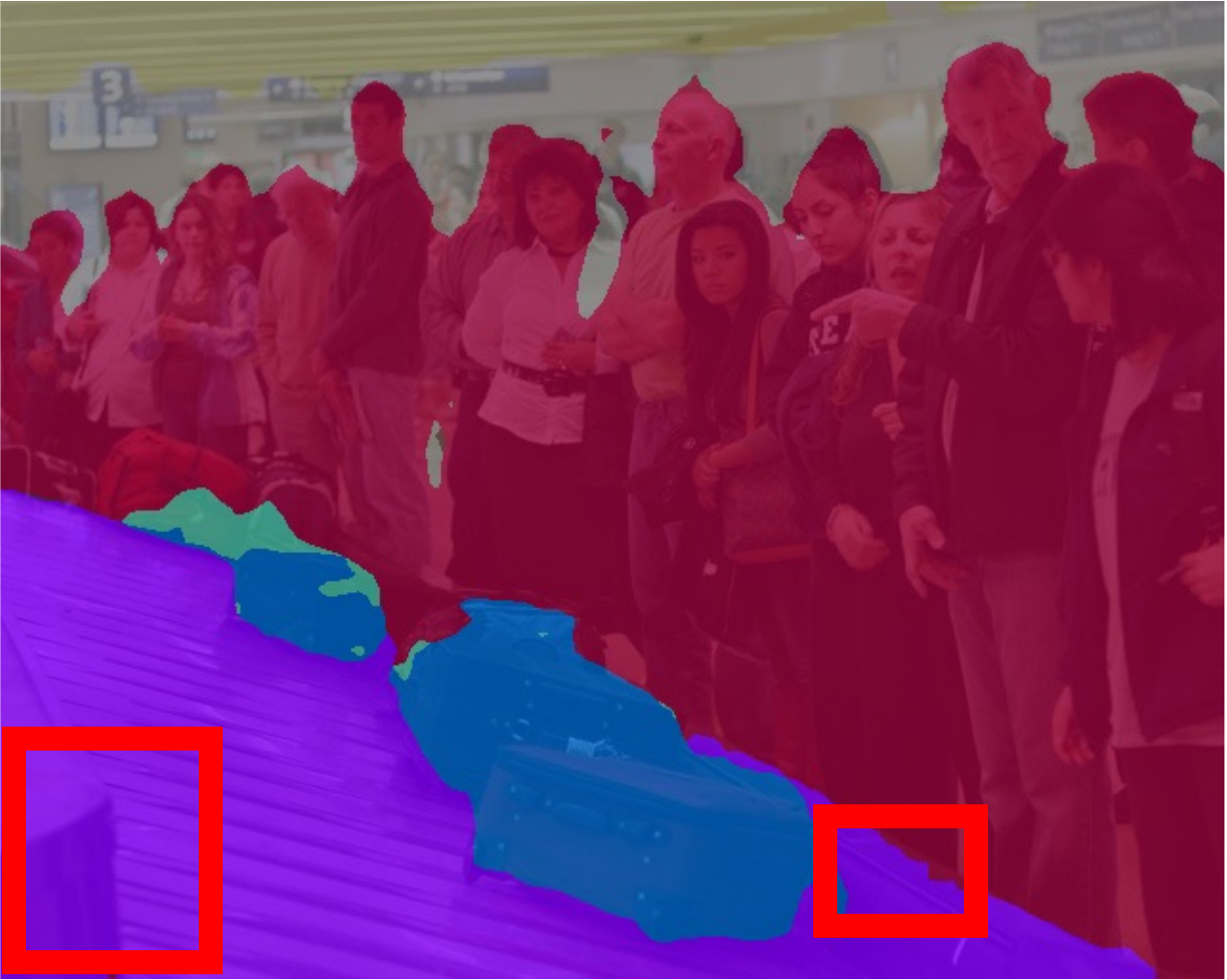} \\ 
  \AddImg{img_00000203.jpg} \hfill \AddImg{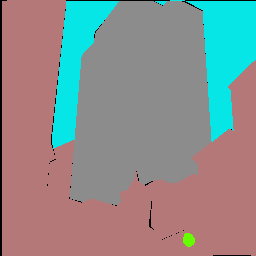} \hfill \AddImg{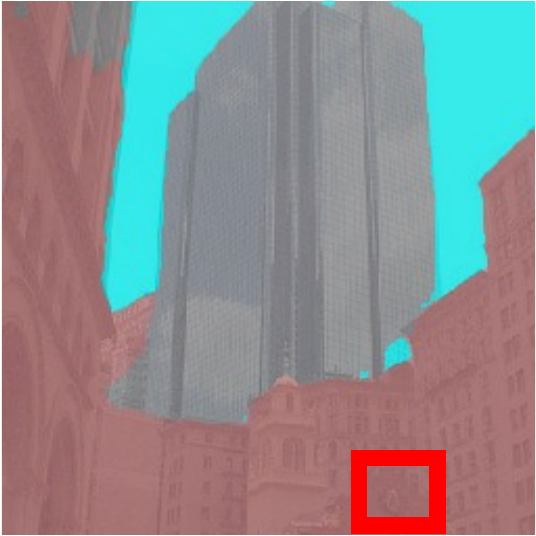} \hfill \AddImg{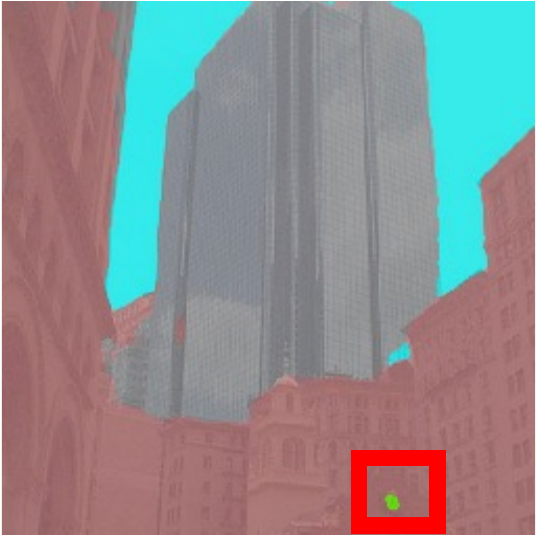} \\ 
 \vspace{-0.06in}
  \leftline{\footnotesize \hspace{0.25in} Image \hspace{0.55in} GT \hspace{0.45in} SegNeXt-T \hspace{0.1in} SegNeXt-T /w Ours }
  \caption{Visualization of the offset learning paradigm on SegNeXt.
  Compared to the baseline SegNeXt-T, applying offset learning paradigm enables the model to segment objects more accurately, especially small objects (\eg, the clock in the third image).
  }
  \vspace{-8pt}
  \label{fig:vis_segnext}
\end{figure}

\myPara{SegFormer with offset learning.}
To systematically evaluate the adaptability of our offset learning paradigm across varying model capacities, we perform comprehensive experiments on six SegFormer architectures (B0-B5) and report results on ADE20K and COCO-Stuff.
To ensure a fair comparison, we employ the results provided by mmsegmentation~\cite{mmseg2020}, wherein the model's FLOPs are smaller than those reported in the paper.
From B0 to B5, the average mIoU improvements on the two datasets are 2.7, 2.2, 1.2, 1.3, 1.1, and 1.0, respectively. 
This gradually diminishing performance gain with increasing model scale aligns with the conclusions drawn from experiments on the SegNeXt model, further validating our hypothesis regarding the misalignment between efficient model features and class representations. 
It also underscores the effectiveness of our offset learning paradigm in addressing the issue of misalignment in the traditional per-pixel classification paradigm.

\myPara{Mask2Former with offset learning.}
We perform experiments on Mask2Former-Tiny, as even the Tiny model already has 47.4M parameters. 
As shown in \tabref{tab:mask2former}, the results demonstrate that integrating our method into the mask classification paradigm improves the performance by 2.6 mIoU with only a 0.2M parameter increase. 
As shown in the comparison in \tabref{tab:formulation}, the mask classification paradigm only adjusts class representations, whereas our offset learning paradigm simultaneously aligns both image features and class representations. 
This demonstrates the effectiveness of our method in achieving efficient and concurrent alignment of image features (per-pixel embeddings) and class representations (mask embeddings).

\begin{table}
\centering
\small
\setlength{\abovecaptionskip}{2pt}
\caption{Performance comparison of Mask2Former~\cite{cheng2022masked} and Mask2Former w/ Offset Learning on ADE20K dataset.}
\tabcolsep=0.2cm
\begin{tabular}{lc|ccc}
\toprule
Method      & Offset       & Params (M) & mIoU \\ \midrule
Mask2Former-Tiny &              & 47.4       & 47.7 \\
\rowcolor[rgb]{ .925,  .925,  .925}
Mask2Former-Tiny & \checkmark   & 47.6       & 50.3\textsubscript{(+2.6)} \\ \bottomrule
\end{tabular}
\label{tab:mask2former}
\end{table}

\subsection{Ablation Study}

\myPara{Ablation analysis on core components.}
To systematically validate the efficacy of each component in our~\ours{} framework, we conduct ablation studies on the ADE20K dataset, as detailed in \tabref{tab:ablation_module}
The baseline model (first row) employs a simple convolutional pixel decoder without FreqFusion~\cite{chen2024frequency} and achieves 40.7 mIoU. 
Introducing FreqFusion alone improves the mIoU score by 1.3.
Based on FreqFusion, combining class offset learning and feature offset learning improves mIoU by 0.9 and 1.5, respectively.
This demonstrates that both branches can independently enhance the model performance, with adaptable features yielding a larger impact than adjustable class representations.
The simultaneous incorporation of both branches achieves the best performance, demonstrating their synergistic effect in jointly aligning image features and class representations.

\myPara{Ablation analysis on the effect of channel number.}
As shown in~\tabref{tab:ablation_channel}, we adopt the baseline w/ FreqFusion only as the base model to verify the effect of channel number on the per-pixel classification paradigm.
When the channel number scales up from 64 to 1024, the results show that increasing feature dimensions in per-pixel classification models yields diminishing returns, and when it reaches 2048, the performance experiences a certain decline.
This reveals that increasing the image feature channels and class representation channels for the per-pixel classification paradigm can improve the expressive ability of the model, as higher-dimensional vectors can represent higher-dimensional spaces.

However, the simple approach of enhancing model performance by increasing the number of channels incurs significant computational overhead and has an upper limit. For instance, at a channel of 1024, the model achieves 43.5 mIoU with 11.1G FLOPs, but further increasing the channel number does not enhance performance.
In contrast, our \ours{}-T with offset learning paradigm achieves a performance with 44.2 mIoU while requiring less than half of the computational resources. 
This comparative analysis further substantiates the efficiency and effectiveness of our proposed method in achieving good segmentation accuracy.

\begin{table}
\centering
\small
\setlength{\abovecaptionskip}{2pt}
\caption{Ablation experiments on different components of the proposed~\ours{}.
FF., CO., and FO. represent FreqFusion, class offset learning, and feature offset learning, respectively.}
\begin{tabular}{lcc|ccc}
\toprule
FF.        &     CO.    &     FO.    & Params (M) & FLOPs (G) & mIoU \\ \midrule
           &            &            & 5.9 & 5.1 & 40.7 \\
\checkmark &            &            & 6.1 & 6.0 & 42.0 \\
\checkmark & \checkmark &            & 6.2 & 5.3 & 42.9 \\
\checkmark &            & \checkmark & 6.2 & 5.3 & 43.5 \\
\rowcolor[rgb]{ .925,  .925,  .925}
\checkmark & \checkmark & \checkmark & 6.2 & 5.3 & 44.2 \\ \bottomrule
\end{tabular}
\label{tab:ablation_module}
\end{table}

\begin{table}
\centering
\small
\setlength{\abovecaptionskip}{2pt}
\caption{Ablation experiments on the effect of channel number of image features and class representations on baseline. The models all belong to the per-pixel classification paradigm, not our offset learning paradigm.}
\tabcolsep=5pt
\begin{tabular}{l|ccccccccc}
\toprule

Channel & 64 &  128 & 256 & 512 & 768 & 1024 & 2048 \\ \midrule
Params (M)  & 6.0 &  6.0 & 6.1 & 6.2 & 6.3 & 6.4 & 6.8 \\
FLOPs (G)  & 4.7 & 5.1  & 6.0 & 7.7 & 9.4 & 11.1 & 17.9 \\
mIoU (\%)  &  41.2 & 41.7 & 42.0 & 42.8 & 42.9 & 43.5 & 43.0 \\
\bottomrule
\end{tabular}
\vspace{-5pt}
\label{tab:ablation_channel}
\end{table}

\section{Conclusions}
\label{sec:conclusion}

In this paper, we analyze the limitations of the per-pixel classification paradigm, specifically the misalignment between image features and class representations. 
To address this issue, we propose the offset learning paradigm, which introduces separate feature offset learning and class offset learning branches to explicitly learn the necessary offsets for aligning image features with their corresponding class representations. 
Building upon this paradigm, we design a series of efficient segmentation networks, named \ours{}, containing three different scales.
As a general segmentation paradigm, we also integrate our offset learning paradigm into three representative segmentation methods, including SegFormer, SegNeXt, and Mask2Former with negligible parameters.
Extensive experiments on four widely used datasets demonstrate the effectiveness and efficiency of our proposed offset learning paradigm.

\section*{Acknowledgment}
This work was partially funded by NSFC (No. 62495061, 62276145), 
National Key Research and Development Project of China (No. 2024YFE0100700),
the Science and Technology Support Program of Tianjin, China (No. 23JCZDJC01050), and A*STAR Career Development Fund (No. C233312006). 
This work was also supported in part by the Supercomputing Center of Nankai University.

{
    \small
    \bibliographystyle{ieee_fullname}
    \bibliography{main}
}

\end{document}